\documentclass{article}

\usepackage{arxiv}
\usepackage{amsmath}
\usepackage{amssymb}
\usepackage{amsfonts}

\usepackage[utf8]{inputenc} % allow utf-8 input
\usepackage[T1]{fontenc}    % use 8-bit T1 fonts
\usepackage{hyperref}       % hyperlinks
\usepackage{url}            % simple URL typesetting
\usepackage{booktabs}       % professional-quality tables
\usepackage{amsfonts}       % blackboard math symbols
\usepackage{nicefrac}       % compact symbols for 1/2, etc.
\usepackage{microtype}      % microtypography
\usepackage{cleveref}       % smart cross-referencing
\usepackage{lipsum}         % Can be removed after putting your text content
\usepackage{graphicx}
\usepackage{natbib}
\usepackage{doi}

\usepackage{graphicx}
\usepackage{url}
\usepackage{multirow}

\newtheorem{definition}{Definition}

\title{An evaluation framework for comparing causal inference models}

\author{ \href{https://orcid.org/0000-0003-1729-4124}{\includegraphics[scale=0.06]{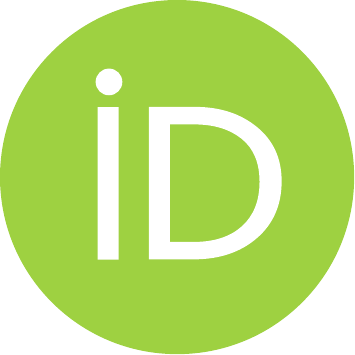}\hspace{1mm}Niki Kiriakidou}\thanks{Corresponding author.} \\
	Department of Informatics and Telematics,\\
	Harokopio University of Athens,\\
	Athens, GR177 78.\\
	\texttt{kiriakidou@hua.gr} \\
	\And
	\href{https://orcid.org/0000-0002-2461-1928}{\includegraphics[scale=0.06]{orcid.pdf}\hspace{1mm}Christos Diou} \\
	Department of Informatics and Telematics,\\
	Harokopio University of Athens,\\
	Athens, GR177 78.\\
	\texttt{cdiou@hua.gr} \\
}

\renewcommand{\shorttitle}{An evaluation framework for comparing causal inference models}

\hypersetup{
pdftitle={An evaluation framework for comparing causal inference models},
pdfsubject={q-bio.NC, q-bio.QM},
pdfauthor={Niki Kiriakidou, Christos Diou},
pdfkeywords={Causal inference, treatment effects, performance profiles, 
	non-para-metric tests, post-hoc tests},
}

\begin{document}
\maketitle

\begin{abstract}
Estimation of causal effects is the core objective of many scientific disciplines. However, it remains a challenging task, especially when the effects are estimated from observational data.
Recently, several promising machine learning models have been proposed for causal effect estimation. The evaluation of these models has been based on the mean values of the error of the Average Treatment Effect (ATE) as well as of the Precision in Estimation of Heterogeneous Effect (PEHE). In this paper, we propose to complement the evaluation of causal inference models using concrete statistical evidence, including the performance profiles of Dolan and Mor{\'e}, as well as non-parametric and post-hoc statistical tests. The main motivation behind this approach is the
elimination of the influence of a small number of instances or simulation on the benchmarking process, which in some cases dominate the results. We use the proposed evaluation methodology to compare several state-of-the-art causal effect estimation models. 
\end{abstract}

% keywords can be removed

\section{Introduction}

Causal inference is a fundamental problem in many scientific areas such as medicine \citep{hofler2005causal}, education \citep{cordero2018causal} and economy \citep{hoover2012economic}. The most effective way to infer the causal effect of a treatment (i.e. intervention)  to an outcome is through a randomized controlled trial. However, in many cases, conducting a randomized controlled trial is not possible, due to financial, ethical or other constraints. Therefore, researchers must determine the effect of treatments by relying on observational data. 

%For example, in the field of medicine, doctors may wish to know a priori the effect of different treatments given to patients, so as to select the most appropriate treatment for every patient. In case a randomized control trial cannot be conducted, we can only rely on observational data to calculate causal effects. 

Observational (as opposed to experimental) data refer to data obtained without any control over independent variables. Researchers simply observe and record the data and do not affect sampling and treatment assignment. In today's big data world, observational data are abundant; nevertheless, using such data to infer causal effects still remains a challenge.

The main obstacle is that for every subject only the factual outcome is observed, i.e. the treatment/outcome combination, which actually took place. The counterfactual outcome is what would have happened, in case we have chosen a different treatment and kept everything else constant. In addition to this problem, treatment assignment is not completely at random, and depends on other factors (e.g., age, socio-economic status or other medical conditions, in the health domain). This results in significant differences between the population that received the treatment (treatment group) and the rest (control group). 

In this paper, we consider the case of a binary treatment $T$. Let $ \mathbf{X}^{n \times m}$ be the design matrix, where $n$ is the number of instances and $m$ is the number of features, $T\in \mathbb\{0,1\}^{n}$ is the treatment assignment and $Y$ is a random variable, with $Y_{0}$ denoting the outcome for a sample when $T = 0$ and $Y_{1}$ the outcome when $T = 1$. 

Let's assume that $\mathbf{x}_{i} \in \mathbf{X}$, $t_{i}\in \mathbb[0,1]$ and $y(\mathbf{x}_{i},t_{i})$ is the outcome of a unit $i = 1,2,\dots,n$. The average treatment effect is defined as follows:
\begin{equation}
	ATE =  \frac{1}{n} \sum_{i=1}^{n} [y(\mathbf{x}_{i},1)-y(\mathbf{x}_{i},0)] 
\end{equation}
It is worth mentioning that, for every unit $i$ in the dataset, only the factual outcome is observed, i.e. $y(\mathbf{x}_{i},1)$ or $y(\mathbf{x}_{i},0)$, which stands for the outcome of each unit in treatment and control group, respectively. As a result, we need to estimate the counterfactual outcomes. The estimated outcomes are denoted as $\hat{y}(\mathbf{x}_{i},1)$ and $\hat{y}(\mathbf{x}_{i},0)$, for treatment and control group, respectively. Then, the estimation of average treatment effect is defined as:
\begin{equation}
	\hat{ATE} =  \frac{1}{n} \sum_{i=1}^{n} [\hat{y}(\mathbf{x}_{i},1)- \hat{y}(\mathbf{x}_{i},0)]
\end{equation}
For the population causal effect we report the absolute error on the average treatment effect,  
\begin{equation}
	\left| \epsilon_{ATE}\right| = \left| \frac{1}{n} \sum_{i=1}^{n} [y(\mathbf{x}_{i},1)-y(\mathbf{x}_{i},0)] - \frac{1}{n} \sum_{i=1}^{n} [\hat{y}(\mathbf{x}_{i},1)- \hat{y}(\mathbf{x}_{i},0)]\right| \label{Eq: error_ATE}
\end{equation}
In order to measure the accuracy of the individual treatment effect estimation, we use the Precision in Estimation of Heterogeneous Effect (PEHE), which is defined as:
\begin{equation}
	\epsilon_{PEHE} =  \frac{1}{n} \sum_{i=1}^{n} [(y(\mathbf{x}_{i},1)-y(\mathbf{x}_{i},0)) -  (\hat{y}(\mathbf{x}_{i},1)- \hat{y}(\mathbf{x}_{i},0))] ^{2}\label{Eq: error_PEHE}
\end{equation}

In the literature, the traditional approach to evaluate the performance of causal inference models
is to compare the average value of $\left| \epsilon_{ATE}\right|$ and/or $\epsilon_{PEHE}$ 
\citep{louizos2017causal,shi2019adapting, shalit2017estimating}. 
However, this approach may provide us misleading results, since all simulations are equally considered
which implies that the difficulty of each simulation of the benchmarking process is not taken into account.
As a result, a small number of the most difficult problems can tend to dominate these results \citep{dolan2002benchmarking,livieris2020improved}.
Additionally, another major drawback is that this approach does not provide 
us any information whether the performance of two or more models is equivalent 
neither quantify the difference between their performance.

In this work, we propose a comprehensive evaluation framework for comparing models for treatment effects and PEHE. The proposed framework is based on evaluating the performance of causal effect estimation models as estimators and as predictors. For providing concrete and empirical evidence, we utilize performance profiles of \cite{dolan2002benchmarking}, as well as non-parametric and post-hoc tests \citep{derrac2011practical}.
The rationale behind our approach is that when evaluating causal effect estimation models using Eqs. \eqref{Eq: error_ATE} and \eqref{Eq: error_PEHE} in multiple experimental evaluations, it is common for a small number of simulations to dominate the results. In these cases, reporting only the average value of the errors may be misleading.

The remainder of this paper is organised as follows: Section 2 presents a
review of neural network-based and tree-based models for the estimation of treatment effects.
Section 3 provides a detailed description of the theoretical framework of performance profiles as well as a complete presentation
of statistical multiple comparison analysis, in order to evaluate the performance of causal inference models. 
Section 4 provides information about the two datasets we used in this work.
Section 5 presents a detailed experimental analysis, focusing on the evaluation of the proposed model and an extended multiple comparisons statistical analysis of models
Section 6 summarizes the main findings and concludes this paper by identifying interesting directions for future work.

\section{Causal inference models}

In the literature, several models have been proposed for the estimation of causal effects. Here, we focus our attention to the most widely used and efficient causal inference models, which can be divided into two categories; neural network-based models and tree-based models.  

\subsection{Neural network-based models}

\cite{shalit2017estimating} proposed the Counterfactual Regression (CFR) framework for estimating individual treatment effects. CFR aims at learning a balanced representation using a prediction model, so that the control and the treatment group distributions look similar. The authors used two integral probability metrics: Wasserstein distance (Wass) \citep{villani2009optimal} and Maximum Mean Discrepancy (MMD) \citep{gretton2012kernel}, in order to measure the distances between two distributions. Furthermore, they introduced a generalization bound for the estimation of individual treatment effect where every individual is only identified by its features. The authors also  proposed the Treatment Agnostic Representation Network (TARNet), which is a variant without balance regularization. 

\cite{shi2019adapting} proposed Dragonnet, which consists of a neural network model for estimating treatment effects from observational data. Dragonnet aims at improving the estimations of average treatment effect and individual treatment effect through the propensity score (i.e., the probability $Pr(T=1\, | \, X=\mathbf{x})$ that a particular sample with covariates $\mathbf{x}$  has received the treatment). Furthermore, the authors proposed a procedure to induce bias based on non-parametric estimation theory to further improve treatment effect estimation. The authors also evaluated Dragonnet to a multi-stage procedure, named NEDnet, which is a neural network with similar architecture with Dragonnet. More specifically, NEDnet first trained using a pure treatment prediction objective. Then, the last layer is removed, and replaced with an outcome-prediction neural network matching the one used by Dragonnet.

\subsection{Tree-based models}

\cite{chipman2010bart} developed a Bayesian ``sum-of-trees'' model named Bayesian Additive Regression Trees (BART). BART model is based on 
a Bayesian non-parametric approach, which fits a parameter rich model by utilizing a strongly influential prior distribution. More specifically, it uses the sum of trees to approximate the average value of the outcomes given a set of covariates, $E[Y|\mathbf{x}]$. The main idea of BART model is to impose a prior, which regularizes the fit by keeping the individual tree effects small in order to elaborate the sum-of-trees model. Additionally, for fitting the sum-of-trees model, BART uses a tailored version of Bayesian backfitting Markov Chain Monte Carlo \citep{hastie2000bayesian}.

\cite{kunzel2019metalearners} proposed a new methodology for predicting treatment effects.
The main idea is to estimate the outcome by using 
all of the features together with the treatment indicator, without proving any special role to the treatment indicator.
In more simple words, the treatment indicator 
is included and treated by the based learner like any other feature.
In the literature, a variety of  base learners such as
Linear Regression \citep{montgomery2021introduction}, Random-Forest \citep{breiman2001random} and $k$-Nearest Neighbors \citep{aha2013lazy} were used providing some interesting results.
However, this approach has two disadvantages; (i) in case the treatment and control groups are very different in covariates, a single model is probably not sufficient to encode the different relevant dimensions and smoothness of features for both groups \citep{alaa2018limits}; (ii) a tree-based base learner
may completely ignore the treatment
assignment by not choosing/splitting on it \citep{kunzel2019metalearners}.

\cite{wager2018estimation} developed a forest-based method for estimating heterogeneous treatment effect. This method consists of an extension of the 
efficient and widely used Random-Forest algorithm \citep{breiman2001random}. 
In more detail, their proposed method, named Causal Forest (C-Forest) is composed by a number of causal trees, which estimates the effect of the treatment at the leaves of the trees. 
A significant advantage of C-Forest 
over the traditional approaches to non-parametric estimation of heterogeneous treatment effects is that the performance of the former is not degradated as the  number of covariates is increasing \citep{wager2018estimation}.

\section{Proposed evaluation framework}

In this section, we present a comprehensive framework for the performance evaluation of the neural network-based and tree-based models. 
%Firstly, we will provide a detailed description of the \textit{performance profiles} of Dolan and Mor{\'e} \cite{dolan2002benchmarking}.
%
In the literature, the traditional way for evaluating the performance of causal inference models is through the average value of the $|\epsilon_{ATE}|$ and $\epsilon_{PEHE}$. 
However, this approach gives us misleading results, since all the number of problems, even the difficult ones, are equally considered for the evaluation of the model.
For this purpose, we use the \textit{performance profiles} of Dolan and Mor{\'e} \citep{dolan2002benchmarking} 
and a \textit{statistical multiple comparison analysis} for evaluating the performance of the models
in order to overcome the problem of the domination of a  small number of the most difficult problems over the results. 

Next, we provide a detailed presentation of
the tools which compose the proposed framework
for evaluating the performance of models to infer causality. 

\subsection{Performance profiles}

\cite{dolan2002benchmarking} presented a methodology for evaluating the effectiveness and robustness of the set of models $M$ on a test set $S$. More specifically, the authors proposed the performance profiles, which provides a wealth of information such as model’s efficiency, robustness and probability of success in compact form \citep{livieris2018performance, livieris2020advanced}. 

In more detail, suppose that there are $n_p$ problems and a set of $n_M$ models for every simulation $s$. Furthermore, let $a_{s,m}$ be $\epsilon_{ATE}$ or $\epsilon_{PEHE}$ by model $m$ for simulation $s$. It is compared the  performance on simulation $s$ by model $m$ with the best performance by any model on this problem, utilizing the performance ratio defined as follows
\begin{equation}
	r_{s, m} =  \frac{a_{s, m}}{min[{a_{s,m} : m\in M}]}
\end{equation}
It is also required to obtain an overall assessment of performance, except of the performance of the model $m$ on a given simulation. Therefore, we are calculating the cumulative distribution function for the performance ratio

\begin{equation} 
	p_{m}(a) = \frac{1}{n_{s}}size[s \in S : r_{s,m} \le \ a],
\end{equation}
where $ a\in \mathbb{R} $ is the factor of the best possible ratio.

Notice that the \textit{performance profile} $p_{m}(a) : \mathbb{R} \rightarrow [0,1] $ for a model is a non-decreasing, piecewise constant function, continuous from the right at each breakpoint \citep{dolan2002benchmarking}. In particular, the performance profile plots the fraction $P$ of problems for which any given model is within a factor $a$  of the best model. This means that the model which outperforms the rest of the models, is the one whose performance profile plot lies on top right.

It is worth highlighting that the use of performance profiles eliminates the influence of a small number of problems on the benchmarking process, which tend to dominate the results \citep{livieris2020improved,livieris2019adaptive}. 
Finally, the vertical axis provides the percentage of the simulations, which were successfully addressed by each model
according to the factor of the best solver (efficiency),
while the horizontal axis summarizes the percentage of the problems for which a model exhibits the best performance (robustness).

\subsection{Statistical multiple comparison analysis}

In the statistical literature, comparison among multiple models is usually carried out by means of non-parametric statistical tests, such as the Friedman test, which constitutes a well-known and widely utilized procedure for  examining the differences between more than two models.

The Friedman test constitutes a non-parametric test,
analogue of the parametric two-way analysis of variance \citep{friedman1940comparison}.
Its main objective is to detect significant differences between the effectiveness of evaluated models,
based on a sample of simulations. An attractive advantage of Friedman test is that the commensurability of the measures across different simulation is not required, since this test is non-parametric \citep{derrac2011practical}.
Furthermore, since it does not assume the normality of the sample means, it is robust to outliers.
This non-parametric test, ranks the scores of each model and uses them in the calculation of the statistic, instead of using the scores themselves. Notice that in case of ties, average ranks are computed. It is worth mentioning
that this test ranks the models from the best to the worst \citep{derrac2011practical,garcia2010advanced}.

Suppose that $r_i^j$ be the rank of the $j$-th of $k$ models on the $i$-th of $M$ problems. 
The Friedman test requires the computation of the average ranks of algorithms $R_j = \frac{1}{n} \sum_i r_i^j$.
Under the null-hypothesis
$
H_0: \{ \text{all models perform similarly with non-significant differences} \}
$,
the Friedman test statistic is calculated by:
\begin{equation}
	F_j = \frac{12n}{k(k+1)} \left[ \sum_j R_j^2 - \frac{k(k+1)^2}{4} \right],
\end{equation}
which is distributed according to a $\chi^2$ distribution with $k-1$ degrees of freedom.

Once Friedman's test rejects $H_0$, we may proceed with a post-hoc
test in order to identify which pair of models differ significantly.
In this research, we employ Bergmann-Hommel  test, 
which was demonstrated as the most powerful test 
for determine the distinctive models in pairwise comparisons \citep{bergmann1988improvements,garcia2008extension}.

For presenting Bergmann-Hommel's procedure, we need the following
definition.

\begin{definition}\label{Def:Exhaustive}
	\textbf{\citep{derrac2011practical}.}
	An index set of hypotheses $I \subseteq \{1, \dots, m\}$ is called
	exhaustive if exactly all $H_j$ with $j\in  I$, could be true.
\end{definition}

Under Definition~\ref{Def:Exhaustive}, the Bergmann-Hommel procedure
rejects any hypothes is $H_j$ with $j$ is not included in the acceptance set $A$
is the index set of null hypotheses, which are retained and it is defined by
\begin{equation}
	A = \bigcup \{I\,:\, I \ \text{exhaustive}, \min \{ P_i \, : \, i \in I\} > a/ |I| \}.
\end{equation}
Additionally, \cite{wright1992adjusted} 
summarized the formula for computation of each Adjusted P-Value (APV), that is 
\begin{equation}
	\text{APV}_i = \min \{v; 1\},
\end{equation}
where $v = \max \{ \|I\| \cdot \min\{ p_j, j\in I \} \, : \, I \ \text{exhaustive}; i\in I \}.$

%\clearpage

\section{Data}

Generally, it is a challenging task to evaluate the performance of a model based on real-world data, due to the nature of the problem on causal inference, since we scarcely have access to the ground truth causal effects.
However, to deal with this difficulty, we count on synthetic and semi-synthetic datasets for the empirical evaluation of causal estimation procedures.

%\subsection{IHDP}

\textbf{IHDP dataset.} The first dataset used for the estimation of individual and population causal effects is the semi-synthetic IHDP dataset, which was introduced by  \cite{hill2011bayesian}.
This dataset was constructed from the Infant Health and Development Program and the
outcome and treatment assignment are fully known. It comprises of 25 features
regarding children and mothers and contains 747 units in which 608 belong to the control group is 608 while the rest 139 belong to the treatment group. Additionally, we studied the effect of home visits by specialists on future cognitive test scores. Finally, we utilized 1000 realizations from the NPCI package \citep{dorie2016npci}.

%\subsection{Synthetic dataset}

\textbf{Synthetic dataset.} This dataset is a toy dataset introduced by \cite{louizos2017causal} and it is generated conditioned on the hidden confounder variable $Z$. More analytically, the process for the generation of synthetic dataset is the following:
\begin{eqnarray*}
z_{i} & \sim & Bern(0.5)\\
t_{i}|z_{i} & \sim & \text{Bern}(0.75z_{i}+0.25(1-z_{i}))\\
x_{i}|z_{i} & \sim & \mathcal{N}(z_{i}, \sigma_{z_{1}}^{2}z_{i} + \sigma_{z_{0}}^{2}(1- z_{i} ))\\
y_{i}|t_{i},z_{i} & \sim & \text{Bern} ( \text{Sigmoid}( 3(z_{i} + 2(2t_{i}-1))) )
\end{eqnarray*}
where the treatment variable $T$ is a mixture of Bernoulli distribution, the proxy to the confounder $X$ is a mixture of Gaussian distribution, the outcome $Y$ is determined as a logistic sigmoid function,  $\sigma_{z_{0}}=3$ and $\sigma_{z_{0}}=5$. This generation process introduces hidden confounding between $T$ and $Y$ as they both depend on the mixture assignment $Z$.

\section{Experimental Analysis}

%NEW TEXT
In this section, we provide a detailed experimental analysis of the performance of neural network-based and tree-based models for IHDP and Synthetic datasets. For both datasets, we present the performance profiles and statistical multiple comparison analysis of the models.

The performance of each model was measured using the metrics 
$|\epsilon_{ATE}|$  and  $\epsilon_{PEHE}$, which are respectively defined by \ref{Eq: error_ATE} and \ref{Eq: error_PEHE}, respectively, as in \citep{shi2019adapting, shalit2017estimating}.
It is worth highlighting, that $|\epsilon_{ATE}|$ metric and $\epsilon_{PEHE}$ are used to compare the
evaluated neural network and tree-based models as estimators and predictors. 

%$$
%\left| \epsilon_{ATE}\right| = \left| \frac{1}{n} \sum_{i=1}^{n} [y_{i}(x,1)-y_{i}(x,0)] - \frac{1}{n} \sum_{i=1}^{n} [\hat{y}_{i}(x,1)- \hat{y}_{i}(x,0)]\right| 
%$$
%and
%$$
%\left| \epsilon_{PEHE}\right| =  \frac{1}{n} \sum_{i=1}^{n} [(y_{i}(x,1)-y_{i}(x,0)) -  (\hat{y}_{i}(x,1)- \hat{y}_{i}(x,0))] ^{2}
%$$

%where $_{i}(x,1)$ and $f_{i}(x,0)$ are the outcomes of a unit $i$ in treated and control group, respectively, while $\hat{f}_{i}(x,1)$, $\hat{f}_{i}(x,0)$ are the corresponding estimations of the outcome. 

The implementation code was written in Python 3.7 using Keras library \citep{gulli2017deep} while
the detailed experimental results for each model regarding both datasets can be found in [\emph{redacted for review}]
%\url{https://github.com/kiriakidou/An-evaluation-framework-for-comparing-causal-inference-models}.

Next, we evaluate the performance of: 

\begin{itemize}
	\item ``{Dragonnet}'', which stands for Dragonnet model of  \cite{shi2019adapting}.
	
	\item ``{TARNet}'', which stands for TARNet model of  \cite{shalit2017estimating}.
	
	\item ``{NEDnet}'', which stands for NEDnet model of  \cite{shi2019adapting}.
	
	\item ``{R-Forest}'', which stands for "S-learner" methodology of  \cite{kunzel2019metalearners} using as base learner Random-Forest \citep{breiman2001random}.
	
	\item ``{C-Forest}'', which stands for C-Forest model of  \cite{wager2018estimation}.
	
	\item ``{BART}'', which stands for BART model of  \cite{chipman2010bart}.
\end{itemize}

All models used the parameters introduced in their original papers. 

\subsection{Results on IHDP dataset}

\begin{figure}[ht]
	\centering
	\includegraphics[width=10cm]{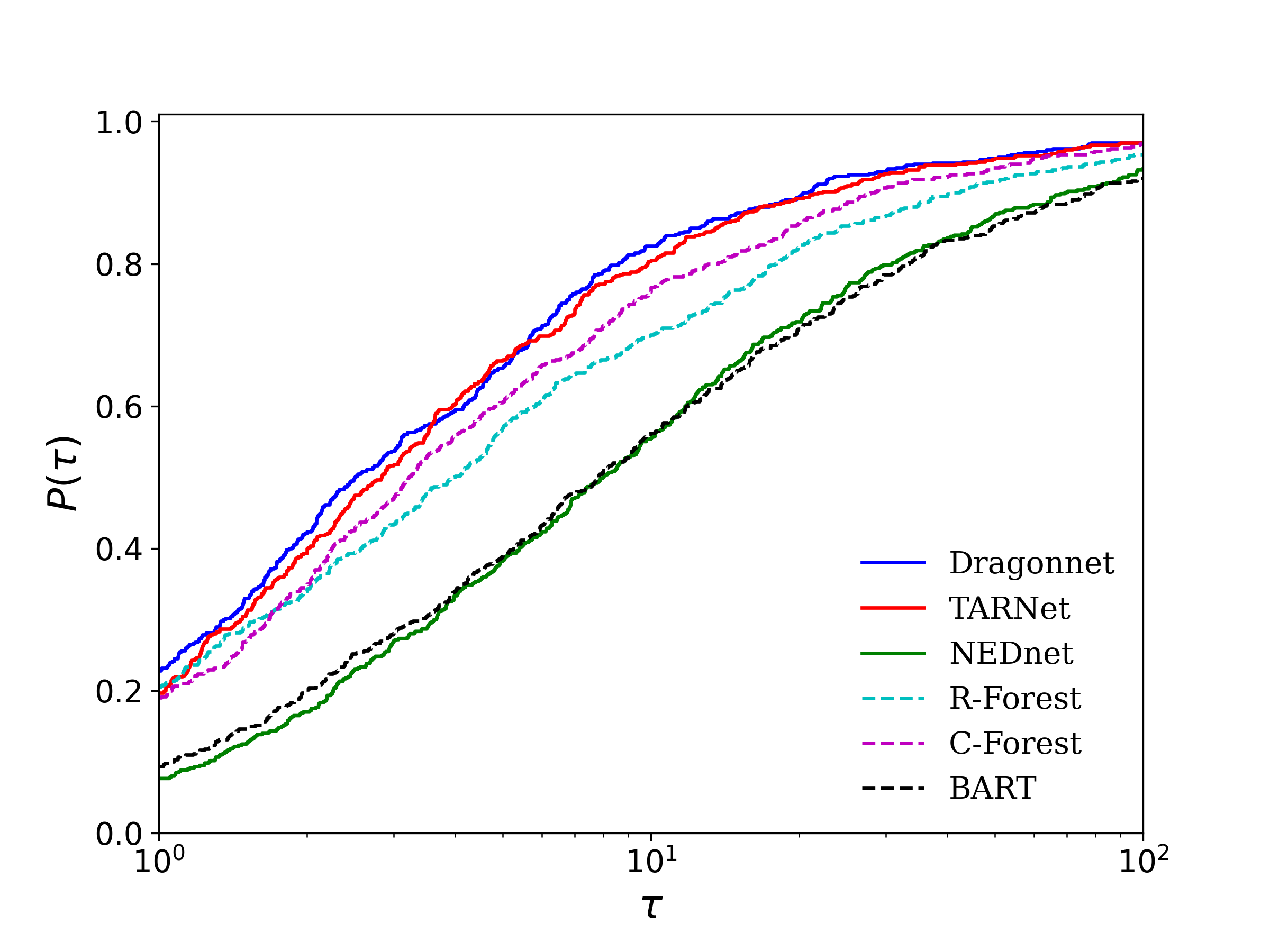}
	\caption{Log$_{10}$ scaled performance profiles based on $|\epsilon_\textbf{ATE}|$}\label{Fig:IHDP_ATE_profile}
\end{figure}

Figure \ref{Fig:IHDP_ATE_profile} presents the performance profiles of neural network-based and tree-based model, based on $\epsilon_{ATE}$ metric. Dragonnet exhibited the best performance in terms of efficiency, slightly outperforming TARNet, R-Forest and C-Forest. More specifically, Dragonnet reported 23\% of simulations with the best (lowest) $\epsilon_{ATE}$, while TARNet, R-Forest and C-Forest presented 20\%, 20\% and 19\%, respectively. In contrast, BART and NEDnet exhibited the worst performance, solving only the 9\% and the 8\% of simulations with the lowest $|\epsilon_{ATE}|$, respectively. Finally, it is worth noticing that Dragonnet and TARNet demonstrated the best performance in terms of robustness, since their curves lie on top.\vspace{-.5cm} 

Table~\ref{Table:IHDP ATE Friedman} presents the Friedman average ranking of all evaluated models, which represent the associated effectiveness of each model.
Notice that the models are ordered from the best (lowest) to the worst (highest) ranking. Clearly, Dragonnet was the best performing model 
followed by TARNet, whereas BART and NEDnet were the worst.
Furthermore, the Friedman statistic $F_f$ with 5 degrees of freedom is equal to 314.33 while the $p$-value is equal to $1.73\cdot 10^{-10}$, which 
suggests the existence of significant differences among the evaluated models.

\begin{table}[!htp]
	\centering 
	\setlength{\tabcolsep}{5pt}
	\renewcommand{\arraystretch}{1}
	\begin{tabular}{cc}
		\toprule
		Algorithm&Ranking\\
		\midrule
		Dragonnet & 2.755\\
		TARNet & 3.016\\
		C-Forest & 3.227\\
		R-Forest & 3.381\\
		BART & 4.245\\
		NEDnet & 4.376\\
		\bottomrule
	\end{tabular}
	\caption{Friedman average rankings of evaluated models based on $|\epsilon_{ATE}|$ for IHDP}\label{Table:IHDP ATE Friedman}
\end{table}

Table \ref{Table:IHDP_ATE_Bergmann} presents the information about the state of rejection 
of all the hypotheses, comparing the models, by summarizing
the APVs with Bergmann Hommel's procedure $p_{Berg}$ with
$\alpha = 5\%$ level of significance for the 15 established comparisons.
Each row contains a hypothesis if the first model
(left side) outperforms the second one (right side), the corresponding
$p_{Berg}$ value and if the hypothesis is rejected or not.
Notice that the hypotheses are ordered from the most to the least significant differences.

In Table \ref{Table:IHDP_ATE_Bergmann} it is worth mentioning that Dragonnet outperforms all tree-based models as well as NEDnet, and has similar performance with TARNet, relative to $|\epsilon_{ATE}|$. TARNet outperforms R-Forest, BART and NEDnet and performed equally well with C-Forest. Furthermore, C-Forest and R-Forest performed similarly and they were only statistically outperformed by Dragonnet. Finally, NEDnet and BART reported the worst performance according both Friedman's and Bergmann's test, since none of them outperforms any other model.

\begin{table}[!ht]
	\centering 
	\setlength{\tabcolsep}{5pt}
	\renewcommand{\arraystretch}{1}
	\begin{tabular}{ccc}
		\toprule
		Hypothesis&$p_{Berg}$\\
		\midrule
		NEDnet vs Dragonnet&0&Rejected\\
		Dragonnet vs BART&0&Rejected\\
		NEDnet vs TARNet&0&Rejected\\
		TARNet vs BART&0&Rejected\\
		NEDnet vs C-Forest&0&Rejected\\
		C-Forest vs BART&0&Rejected\\
		R-Forest vs NEDnet&0&Rejected\\
		R-Forest vs BART&0&Rejected\\
		R-Forest vs Dragonnet&0.000001&Rejected\\
		Dragonnet vs C-Forest&0.000265&Rejected\\
		R-Forest vs TARNet&0.008147&Rejected\\
		Dragonnet vs TARNet&0.082183&Failed to be rejected\\
		TARNet vs C-Forest&0.149083&Failed to be rejected\\
		R-Forest vs C-Forest&0.386149&Failed to be rejected\\
		NEDnet vs BART&0.386149&Failed to be rejected\\
		\bottomrule
	\end{tabular}
	\caption{Multiple comparison test: Bergmann-Hommel's APVs based on $|\epsilon_{ATE}|$ for IHDP}\label{Table:IHDP_ATE_Bergmann}
\end{table}

\begin{figure}[ht]
	\centering
	\includegraphics[width=12cm]{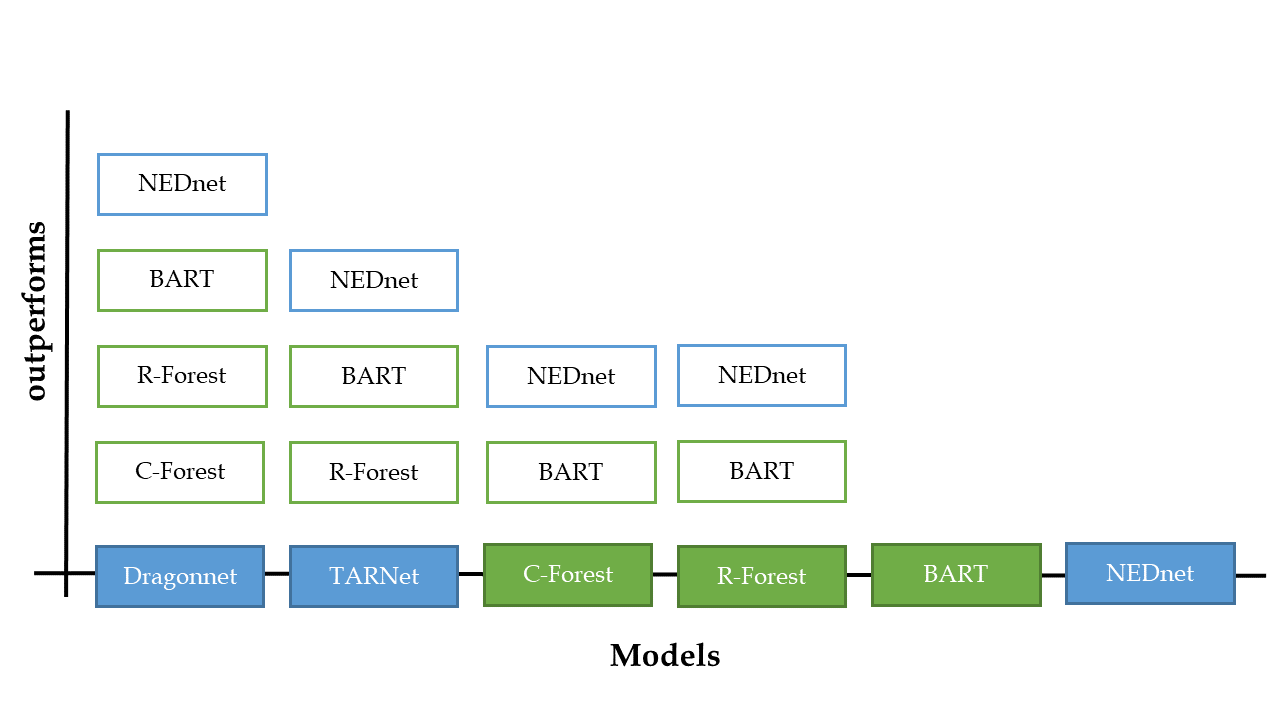}
	\caption{Conclusions for comparison based on $|\epsilon_\textbf{ATE}|$ for IHDP}\label{Fig:Comparison_IHDP_ATE}
\end{figure}

Figure \ref{Fig:Comparison_IHDP_ATE} summarizes the the conducted findings and conclusions of Table \ref{Table:IHDP_ATE_Bergmann} and Table \ref{Table:IHDP_ATE_Bergmann}. More specifically, $x$-axis presents the models based on Friedman ranking. For each of them, y-axis collects the models  which were statistically outperformed according to the Bergmann–Hommel test.\vspace{-.5cm}

\begin{figure}[ht]
	\centering
	\includegraphics[width=10cm]{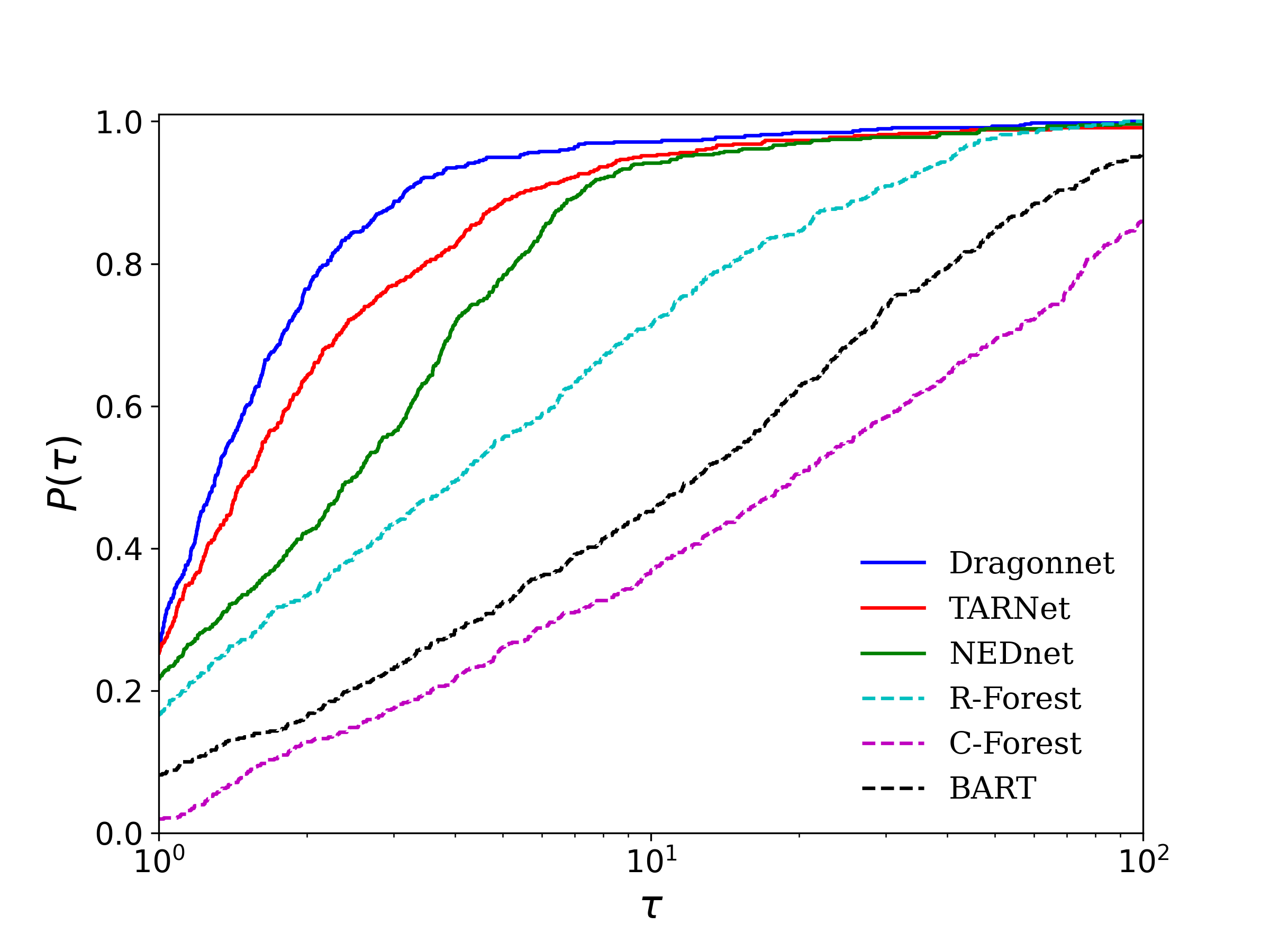}
	\caption{Log$_{10}$ scaled performance profiles based on $\epsilon_\textbf{PEHE}$}\label{Fig:IHDP_PEHE_profile}
\end{figure}

Figure \ref{Fig:IHDP_PEHE_profile} presents the performance profiles of the selected models, based on $\epsilon_{PEHE}$ metric. In terms of efficiency, Dragonnet and TARNet presented the best performance, followed by NEDnet and R-Forest. In more detail,  Dragonnet and TARNet reported 26\% and 25\% with the lowest $\epsilon_{PEHE}$, respectively, while NEDnet and R-Forest exhibited 22\% and 17\%, respectively. Additionally, BART and C-Forest presented the lowest performances solving only the 8\% and 2\% of simulations, respectively. In terms of robustness, Dragonnet illustrated the top curve.

Table~\ref{Table:IHDP PEHE Friedman} presents the Friedman average ranking of all evaluated models, which represent the associated effectiveness of each model.
Dragonnet was the best performing model followed by TARNet, while
C-Forest was the worst.
In addition, the Friedman statistic $F_f$ with 5 degrees of freedom is equal to 1029.74 while the $p$-value is equal to 0, which strongly suggests the existence of significant differences between the evaluated models.

\clearpage

\begin{table}[!htp]
	\centering 
	\setlength{\tabcolsep}{5pt}
	\renewcommand{\arraystretch}{1}
	\begin{tabular}{cc}
		\toprule
		Algorithm&Ranking\\
		\midrule
		Dragonnet & 2.098\\
		TARNet & 2.782\\
		R-Forest & 3.150\\
		NEDnet & 3.166\\
		BART & 4.372\\
		C-Forest & 5.432\\
		\bottomrule
	\end{tabular}
	\caption{Friedman average rankings of evaluated models based on $\epsilon_{PEHE}$ for IHDP}\label{Table:IHDP PEHE Friedman}
\end{table}

\begin{figure}[ht]
	\centering
	\includegraphics[width=12cm]{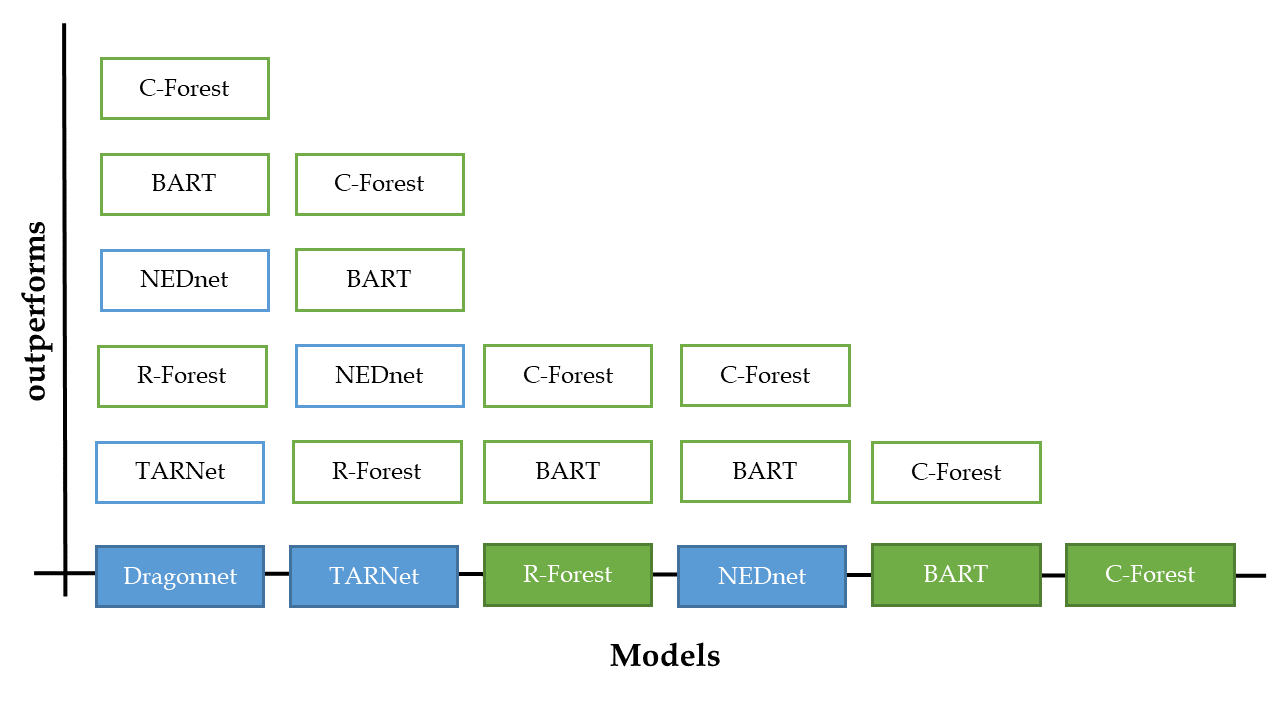}
	\caption{Conclusions for comparison based on $\epsilon_\textbf{PEHE}$ for IHDP}\label{Fig:Comparison_IHDP_PEHE}
\end{figure}

Table \ref{Table:IHDP PEHE Bergmann} suggests that Dragonnet outperformed all other models, while TARNet outperformed NEDnet and all tree-based models. R-Forest and NEDnet had no statistically important differences in their performance and both outperformed C-Forest and BART. Eventually, BART and C-Forest exhibited the worst performance in terms of $\epsilon_{PEHE}$. A graphical overview of the conducted findings of the statistical analysis is presented in Figure \ref{Fig:Comparison_IHDP_PEHE}.

\begin{table}[!ht]
	\centering 
	\setlength{\tabcolsep}{5pt}
	\renewcommand{\arraystretch}{1}
	\begin{tabular}{ccc}
		\toprule
		Hypothesis&$p_{Berg}$\\
		\midrule
		Dragonnet vs C-Forest&0&Rejected\\
		TARNet vs C-Forest&0&Rejected\\
		R-Forest vs C-Forest&0&Rejected\\
		Dragonnet vs BART&0&Rejected\\
		NEDnet vs C-Forest&0&Rejected\\
		TARNet vs BART&0&Rejected\\
		R-Forest vs BART&0&Rejected\\
		NEDnet vs BART&0&Rejected\\
		NEDnet vs Dragonnet&0&Rejected\\
		C-Forest vs BART&0&Rejected\\
		R-Forest vs Dragonnet&0&Rejected\\
		Dragonnet vs TARNet&0&Rejected\\
		NEDnet vs TARNet&0.003519&Rejected\\
		R-Forest vs TARNet&0.003519&Rejected\\
		R-Forest vs NEDnet&0.892434&Failed to be rejected\\
		\bottomrule
	\end{tabular}
	\caption{Multiple comparison test: Bergmann-Hommel's APVs based on $\epsilon_{PEHE}$ for IHDP}\label{Table:IHDP PEHE Bergmann}
\end{table}

Summarizing, by taking into consideration both the performance profiles and statistical analysis, we are able to conclude that Dragonnet and TARNet outperformed the rest models in terms of $|\epsilon_{ATE}|$ and  $\epsilon_{PEHE}$. This suggests that they reported the best performance, as estimators and predictors. Finally, it is worth mentioning, that our experimental analysis illustrated that Dragonnet outperformed TARNet in terms of robustness for both metrics.

%\clearpage

\subsection{Results on Synthetic dataset}

\begin{figure}[ht]
	\centering
	\includegraphics[width=10cm]{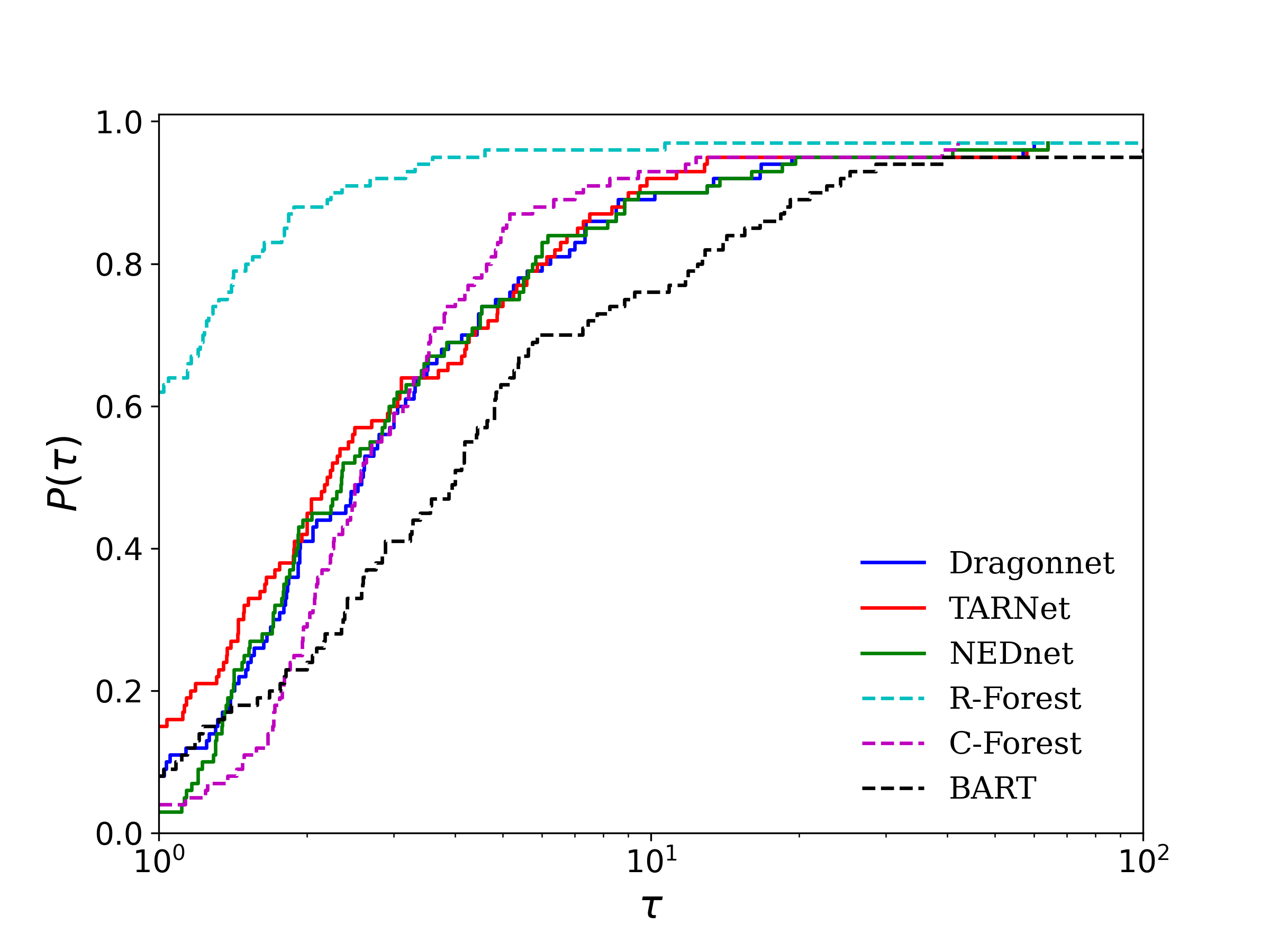}
	\caption{Log$_{10}$ scaled performance profiles based on $|\epsilon_\textbf{ATE}|$}\label{Fig:Synthetic_ATE_profile}
\end{figure}

Figure \ref{Fig:Synthetic_ATE_profile} presents the performance profiles of tree-based models and neural network models, based on metric $|\epsilon_{ATE}|$. It is worth mentioning that R-Forest model, reported the best performance in terms of efficiency and robustness, illustrating the top curve. More specifically, R-Forest outperforms the rest of neural network-based models and tree-based models, exhibiting 62\% of simulations with the lowest $|\epsilon_{ATE}|$. Neural network- based models TARNet, Dragonnet and NEDnet presented 15\%, 8\% and 3\% respectively, with the lowest $|\epsilon_{ATE}|$ score, while tree-based models BART and C-Forest reported poor 
performance, solving only 8\% and 4\% of the simulations, respectively.  
\vspace{-.5cm}

\clearpage

Table~\ref{Table:Synthetic ATE Friedman} presents the Friedman average ranking of all evaluated models, which represents the associated effectiveness of each model.
Clearly, R-Forest was the best performing model, outperforming all neural network-based models, while BART and C-Forest reported the worst performance.
Furthermore, the Friedman statistic $F_f$ with 5 degrees of freedom is equal to 174.25 while the $p$-value is equal to $8.6\cdot 10^{-11}$, which 
suggests the existence of differences among the evaluated models.

\begin{table}[!htp]
	\centering 
	\setlength{\tabcolsep}{5pt}
	\renewcommand{\arraystretch}{1}
	\begin{tabular}{cc}
		\toprule
		Algorithm&Ranking\\
		\midrule
		R-Forest & 1.665\\
		TARNet & 3.13\\
		NEDnet & 3.49\\
		Dragonnet & 3.685\\
		C-Forest & 4.015\\
		BART & 5.015\\
		\bottomrule
	\end{tabular}
	\caption{Friedman average rankings of evaluated models based on $|\epsilon_{ATE}|$ for Synthetic Dataset}\label{Table:Synthetic ATE Friedman}
\end{table}

Table~\ref{Table:Synthetic ATE Bergmann} presents the information about the state of rejection 
of all the hypotheses, comparing the models, by summarizing
the APVs with Bergmann Hommel's procedure $p_{Berg}$ with
$\alpha = 5\%$ level of significance for the 15 established comparisons.

\begin{figure}[ht]
	\centering
	\includegraphics[width=12cm]{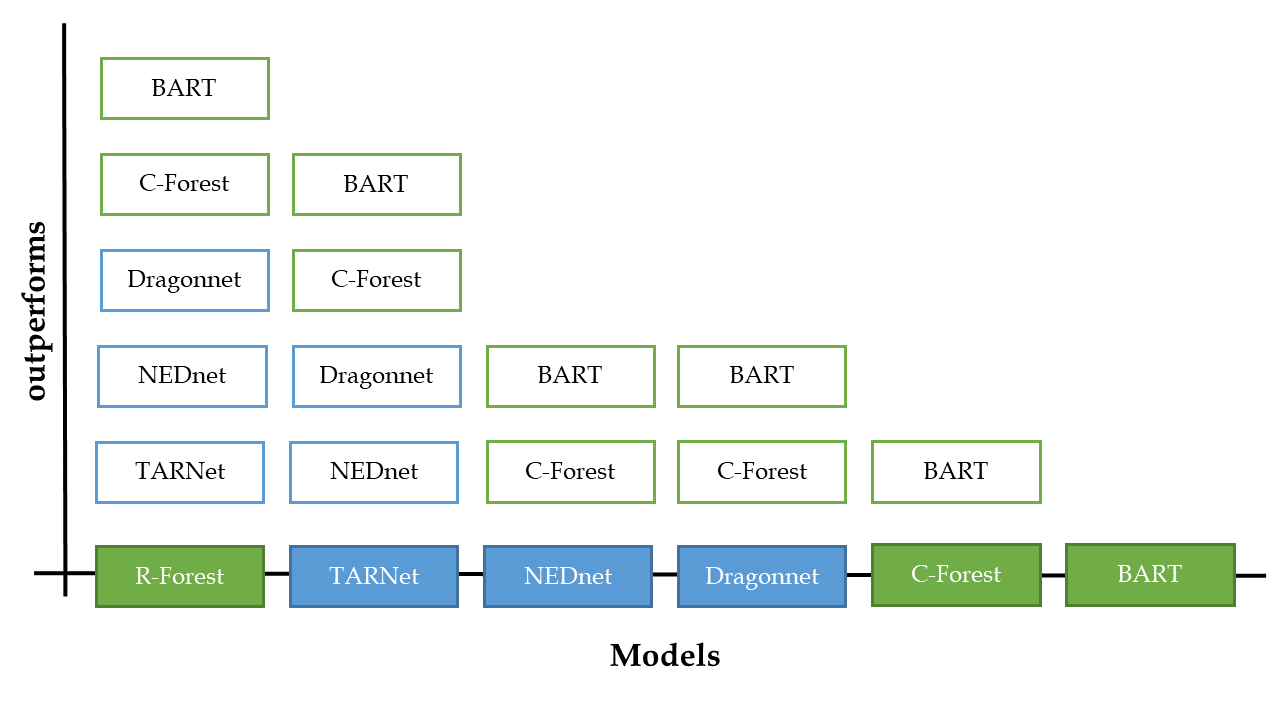}
	\caption{Conclusions for comparison based on $|\epsilon_\textbf{ATE}|$ for Synthetic}\label{Fig:Comparison_Synthetic_ATE}
\end{figure}

More specifically, Table \ref{Table:Synthetic ATE Bergmann} reveals that R-Forest outperforms all of the tree-based and neural network-based models, while TARNet outperformed the rest of the neural network models as well as C-Forest and BART. Furthermore, NEDnet and Dragonnet have equally well performances, and they both outperformed C-Forest and BART. Finally, C-Forest and BART reported the worst performances. Figure \ref{Fig:Comparison_Synthetic_ATE} summarizes and presents in a compact  graphical overview the conclusions extracted from Friedman and Bergmann-Hommel’s test.

\begin{table}[!ht]
	\centering 
	\setlength{\tabcolsep}{5pt}
	\renewcommand{\arraystretch}{1}
	\begin{tabular}{ccc}
		\toprule
		Hypothesis&$p_{Berg}$\\
		\midrule
		R-Forest vs C-Forest&0&Rejected\\
		R-Forest vs Dragonnet&0&Rejected\\
		BART vs TARNet&0&Rejected\\
		R-Forest vs NEDnet&0&Rejected\\
		NEDnet vs BART&0&Rejected\\
		R-Forest vs TARNet&0&Rejected\\
		BART vs Dragonnet&0.000002&Rejected\\
		BART vs C-Forest&0.000628&Rejected\\
		TARNet vs C-Forest&0.004937&Rejected\\
		Dragonnet vs TARNet&0.107794&Fail to be rejected\\
		NEDnet vs C-Forest&0.141663&Fail to be rejected\\
		NEDnet vs TARNet&0.347235&Fail to be rejected\\
		Dragonnet vs C-Forest&0.347235&Fail to be rejected\\
		NEDnet vs Dragonnet&0.461104&Fail to be rejected\\
		\bottomrule
	\end{tabular}
	\caption{Multiple comparison test: Bergmann-Hommel's APVs based on $|\epsilon_{ATE}|$ for Synthetic Dataset}\label{Table:Synthetic ATE Bergmann}
\end{table}

In Figure \ref{Fig:Synthetic_PEHE_profile } it is worth mentioning, that in terms of $\epsilon_{PEHE}$ in Synthetic dataset, C-Forest is solving an extremely high percentage of simulations. More specifically, C-Forest reported 80\% of the simulations. Simultaneously, BART exhibited 30\% of simulations in the same situation. Additionally, the most noticeable result is that R-Forest and all of neural network-based models, were not able to solve any of the simulations with the best $\epsilon_{PEHE}$ and performed poorly since their curves lie on the bottom. Finally, the interpretation of Figure\label{Fig:Synthetic_PEHE_profile} reveals that all the neural network-based models were the worst in terms of robustness.

\begin{figure}[ht]
	\centering
	\includegraphics[width=10cm]{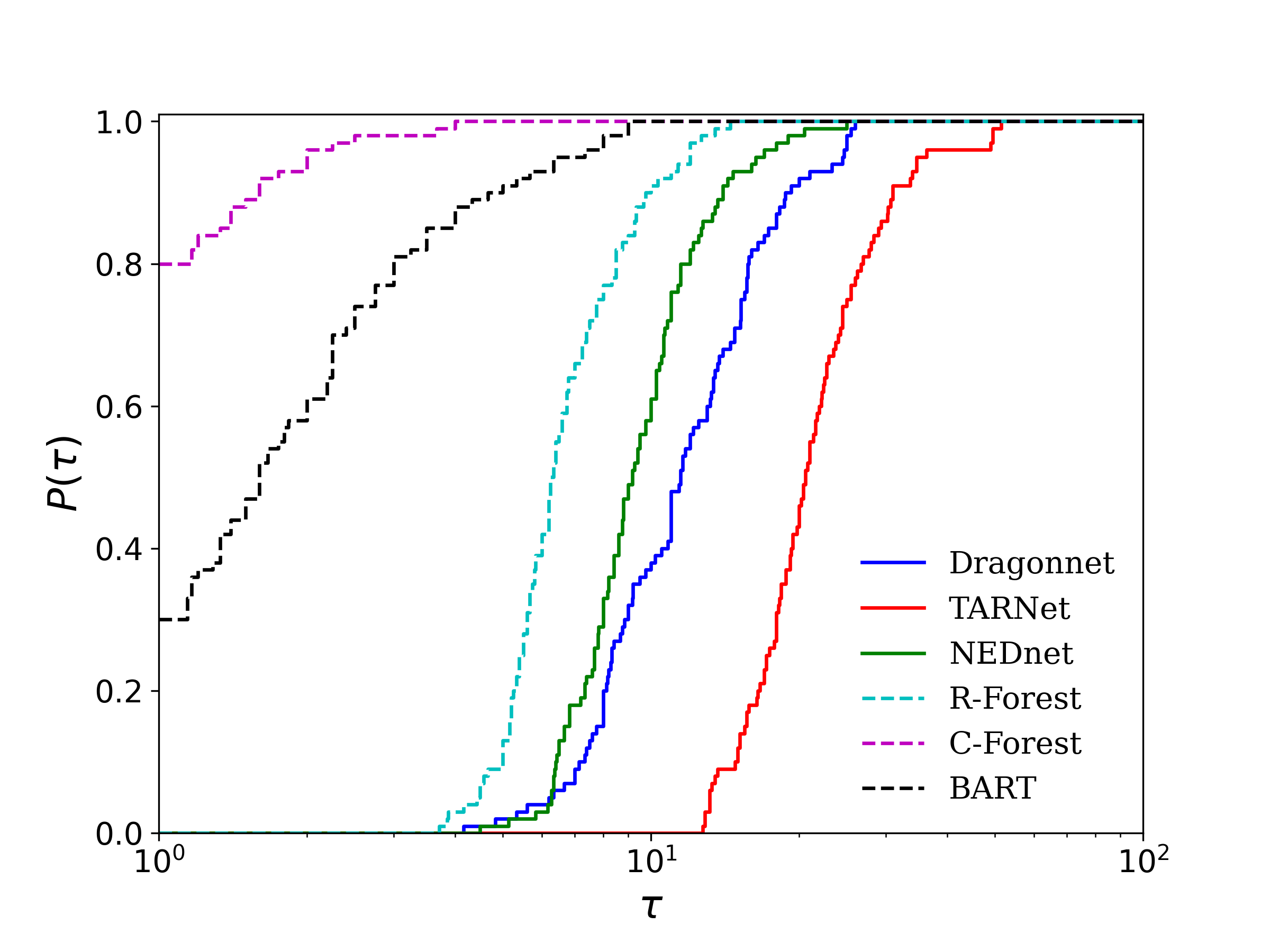}
	\caption{Log$_{10}$ scaled performance profiles based on $\epsilon_\textbf{PEHE}$}\label{Fig:Synthetic_PEHE_profile }
\end{figure}

Table~\ref{Table:Synthetic PEHE Friedman} presents the Friedman average ranking of all evaluated models, which represent the associated effectiveness of each model.
C-Forest was the best performing model followed by BART while 
the neural network-based models were the worst.
In addition, the Friedman statistic $F_f$ with 5 degrees of freedom is equal to 469.41 while the $p$-value is equal to $1.76\cdot 10^{-10}$, which 
suggests the existence of significant differences between the evaluated models.\\ \\

\begin{table}[!htp]
	\centering 
	\setlength{\tabcolsep}{5pt}
	\renewcommand{\arraystretch}{1}
	\begin{tabular}{cc}
		\toprule
		Algorithm&Ranking\\
		\midrule
		C-Forest & 1.25\\
		BART & 1.75\\
		R-Forest & 3.12\\
		NEDnet & 4.15\\
		Dragonnet & 4.74\\
		TARNet & 5.99\\
		\bottomrule
	\end{tabular}
	\caption{Friedman average rankings of evaluated models based on $\epsilon_{PEHE}$ for Synthetic Dataset}\label{Table:Synthetic PEHE Friedman}
\end{table}

Table~\ref{Table:Synthetic PEHE Bergmann} summarizes the state of rejection of all conducted hypotheses, between the evaluated models, by reporting the APVs with Bergmann Hommel's procedure $p_{Berg}$ with $\alpha = 5\%$ level of significance.

\begin{figure}[ht]
	\centering
	\includegraphics[width=12cm]{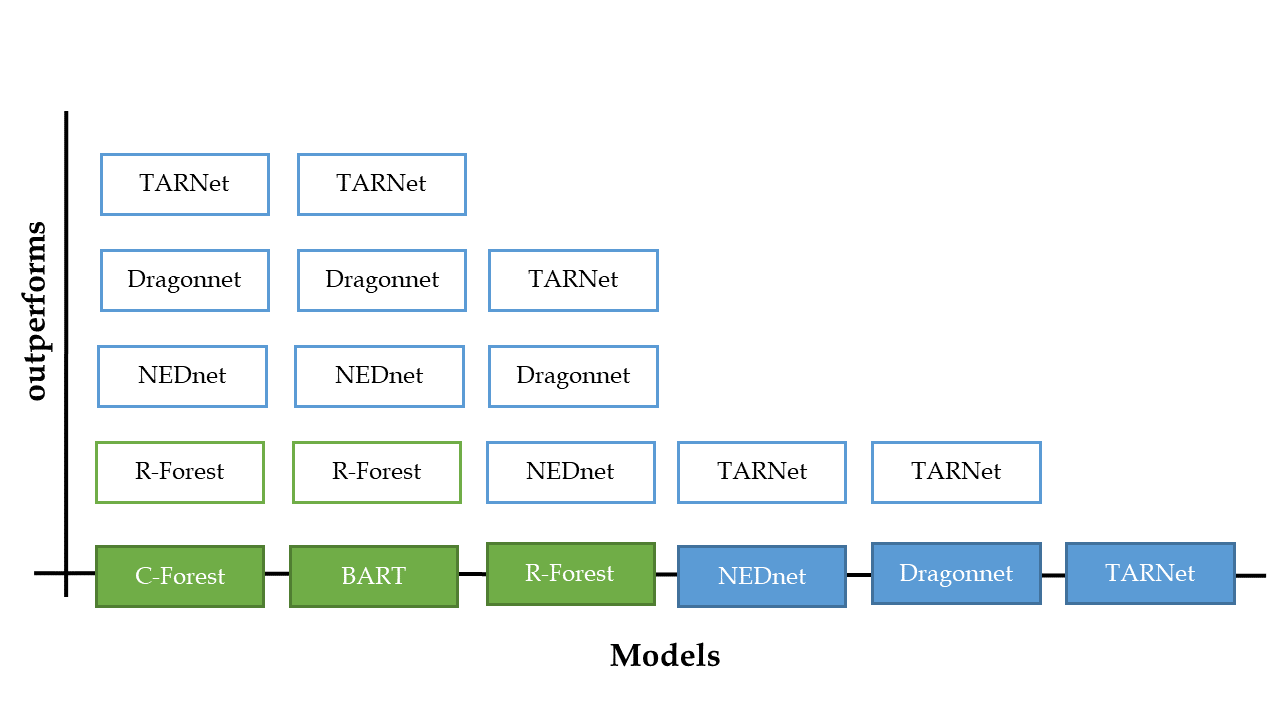}
	\caption{Conclusions for comparison based on $\epsilon_\textbf{PEHE}$ for Synthetic}\label{Fig:Comparison_Synthetic_PEHE}
\end{figure}

In more detail, Table \ref{Table:Synthetic PEHE Bergmann} reveals that both C-Forest and BART exhibited similar performances, simultaneously outperformed all neural network models and R-Forest. Additionally, R-Forest outperformed all neural network models, while NEDnet and Dragonnet outperformed TARNet. The worst performance was held by TARNet. A visual outline of the conducted findings of statistical analysis is reported in Figure \ref{Fig:Comparison_Synthetic_PEHE}.

\begin{table}[!ht]
	\centering 
	\setlength{\tabcolsep}{5pt}
	\renewcommand{\arraystretch}{1}
	\begin{tabular}{ccc}
		\toprule
		Hypothesis&$p_{Berg}$\\
		\midrule
		TARNet vs C-Forest&0&Rejected\\
		BART vs TARNet&0&Rejected\\
		Dragonnet vs C-Forest&0&Rejected\\
		BART vs Dragonnet&0&Rejected\\
		NEDnet vs C-Forest&0&Rejected\\
		R-Forest vs TARNet&0&Rejected\\
		NEDnet vs BART&0&Rejected\\
		R-Forest vs C-Forest&0&Rejected\\
		NEDnet vs TARNet&0&Rejected\\
		R-Forest vs Dragonnet&0&Rejected\\
		R-Forest vs BART&0&Rejected\\
		Dragonnet vs TARNet&0.000007&Rejected\\
		R-Forest vs NEDnet&0.000198&Rejected\\
		NEDnet vs Dragonnet&0.051496&Failed to be rejected\\
		BART vs C-Forest&0.058782&Failed to be rejected\\
		\bottomrule
	\end{tabular}
	\caption{Multiple comparison test: Bergmann-Hommel's APVs based on $\epsilon_{PEHE}$ for Synthetic Dataset}\label{Table:Synthetic PEHE Bergmann}
\end{table}

\vspace{-.5cm}

Summarizing all the above, by taking into consideration both the statistical analysis and performance profiles, it is clear that tree-based model exhibited better performance than neural network models on Synthetic dataset as regards to  $|\epsilon_{ATE}|$ and  $\epsilon_{PEHE}$. This reveals that tree-based model exhibited the best performance as estimators and predictors.

%\clearpage

\section{Conclusions} 

In this work, we proposed a complete framework, for the evaluation of causal inference models in order to estimate treatment effects.
More specifically, we presented an evaluation framework based on the performance profiles of Dolan and Mor{\'e} as well as on multiple statistical analysis for providing strong and concrete statistical evidence. 
In the literature, the traditional approach for the evaluation of these models has been based
on the mean values of the $|\epsilon_{ATE}|$  and $\epsilon_{PEHE}$.
The major disadvantage of this approach is that 
all simulations are considered equal which leads to 
misleading and ambiguous conclusions about the performance of each compared model.
This constitutes the main motivation behind our approach
focusing on eliminating the influence of a small number of simulations on the benchmarking process, which in some cases dominate the results.

In our experiments, we evaluated the performance of neural network-based and tree-based models using two different datasets (IHDP and Synthetic) with different characteristics. 
The experimental analysis based on the proposed evaluation framework revealed that neural-based models presented the best performance on IHDP, while tree-based model exhibited top performance on Synthetic. Therefore, we are able to conclude that no single class of models is the dominant for both datasets. 
Additionally, based on the performance of each single model, we are able to conduct similar 
conclusions (i.e no single model performs well on all problems.). 
In more detail, our analysis showed that for IHDP dataset, Dragonnet and TARNet presented the best results over all models, both as estimators and as predictors. In case of the Synthetic dataset, R-Forest performed better as estimator in terms of $|\epsilon_{ATE}|$, while C-Forest performed better as predictor in terms of $\epsilon_{PEHE}$.

It is worth mentioning that based on the traditional evaluation approach (i.e comparing the mean $|\epsilon_{ATE}|$ and $\epsilon_{PEHE}$ over all simulations), we include similar conclusions regarding the
models' ranking.
However, this approach is misleading in case the variance over all simulations is high
and it does not provide us any information whether the performance of two or more models is equivalent neither quantify the difference between their performance.
In conclusion, we point out that the proposed evaluation framework is able to provide a deep insight to the performance of causal inference models.

In our future work, we intend to apply the proposed evaluation framework including more real-world and synthetic datasets as well as more causal inference models for the estimation of individual and average treatment effect. 
Our aim is to identify the models or the class of models, which is ``dominated'' for a class of problems with specific characteristics. 
This could provide more useful information about the performance as well as the advantages and disadvantages of each compared model.
Finally, it is worth mentioning that our objective and expectation is that this work could be utilized as a reference for  evaluating causal inference models, by offering strong statistical evidence for model's evaluation.

\section*{Aknowledgements} 

The work leading to these results has received funding from the European Union's Horizon 2020 research and innovation programme under Grant Agreement No. 965231, project REBECCA (REsearch on BrEast Cancer induced chronic conditions supported by Causal Analysis of multi-source data)

%\bibliographystyle{unsrtnat}
%\bibliography{references} 

\begin{thebibliography}{30}
	\providecommand{\natexlab}[1]{#1}
	\providecommand{\url}[1]{\texttt{#1}}
	\expandafter\ifx\csname urlstyle\endcsname\relax
	\providecommand{\doi}[1]{doi: #1}\else
	\providecommand{\doi}{doi: \begingroup \urlstyle{rm}\Url}\fi
	
	\bibitem[H{\"o}fler(2005)]{hofler2005causal}
	Marc H{\"o}fler.
	\newblock Causal inference based on counterfactuals.
	\newblock \emph{BMC medical research methodology}, 5\penalty0 (1):\penalty0
	1--12, 2005.
	
	\bibitem[Cordero et~al.(2018)Cordero, Crist{\'o}bal, and
	Sant{\'\i}n]{cordero2018causal}
	Jos{\'e}~M Cordero, V{\'\i}ctor Crist{\'o}bal, and Daniel Sant{\'\i}n.
	\newblock Causal inference on education policies: A survey of empirical studies
	using pisa, timss and pirls.
	\newblock \emph{Journal of Economic Surveys}, 32\penalty0 (3):\penalty0
	878--915, 2018.
	
	\bibitem[Hoover(2012)]{hoover2012economic}
	Kevin~D Hoover.
	\newblock Economic theory and causal inference.
	\newblock \emph{Philosophy of economics}, 13:\penalty0 89--113, 2012.
	
	\bibitem[Louizos et~al.(2017)Louizos, Shalit, Mooij, Sontag, Zemel, and
	Welling]{louizos2017causal}
	Christos Louizos, Uri Shalit, Joris~M Mooij, David Sontag, Richard Zemel, and
	Max Welling.
	\newblock Causal effect inference with deep latent-variable models.
	\newblock \emph{Advances in neural information processing systems}, 30, 2017.
	
	\bibitem[Shi et~al.(2019)Shi, Blei, and Veitch]{shi2019adapting}
	Claudia Shi, David Blei, and Victor Veitch.
	\newblock Adapting neural networks for the estimation of treatment effects.
	\newblock In \emph{Advances in neural information processing systems},
	volume~32, 2019.
	
	\bibitem[Shalit et~al.(2017)Shalit, Johansson, and
	Sontag]{shalit2017estimating}
	Uri Shalit, Fredrik~D Johansson, and David Sontag.
	\newblock Estimating individual treatment effect: generalization bounds and
	algorithms.
	\newblock In \emph{International Conference on Machine Learning}, pages
	3076--3085. PMLR, 2017.
	
	\bibitem[Dolan and Mor{\'e}(2002)]{dolan2002benchmarking}
	Elizabeth~D Dolan and Jorge~J Mor{\'e}.
	\newblock Benchmarking optimization software with performance profiles.
	\newblock \emph{Mathematical programming}, 91\penalty0 (2):\penalty0 201--213,
	2002.
	
	\bibitem[Livieris and Pintelas(2020)]{livieris2020improved}
	Ioannis~E Livieris and Panagiotis Pintelas.
	\newblock An improved weight-constrained neural network training algorithm.
	\newblock \emph{Neural Computing and Applications}, 32\penalty0 (9):\penalty0
	4177--4185, 2020.
	
	\bibitem[Derrac et~al.(2011)Derrac, Garc{\'\i}a, Molina, and
	Herrera]{derrac2011practical}
	Joaqu{\'\i}n Derrac, Salvador Garc{\'\i}a, Daniel Molina, and Francisco
	Herrera.
	\newblock A practical tutorial on the use of nonparametric statistical tests as
	a methodology for comparing evolutionary and swarm intelligence algorithms.
	\newblock \emph{Swarm and Evolutionary Computation}, 1\penalty0 (1):\penalty0
	3--18, 2011.
	
	\bibitem[Villani(2009)]{villani2009optimal}
	C{\'e}dric Villani.
	\newblock \emph{Optimal transport: old and new}, volume 338.
	\newblock Springer, 2009.
	
	\bibitem[Gretton et~al.(2012)Gretton, Borgwardt, Rasch, Sch{\"o}lkopf, and
	Smola]{gretton2012kernel}
	Arthur Gretton, Karsten~M Borgwardt, Malte~J Rasch, Bernhard Sch{\"o}lkopf, and
	Alexander Smola.
	\newblock A kernel two-sample test.
	\newblock \emph{The Journal of Machine Learning Research}, 13\penalty0
	(1):\penalty0 723--773, 2012.
	
	\bibitem[Chipman et~al.(2010)Chipman, George, and McCulloch]{chipman2010bart}
	Hugh~A Chipman, Edward~I George, and Robert~E McCulloch.
	\newblock Bart: Bayesian additive regression trees.
	\newblock \emph{The Annals of Applied Statistics}, 4\penalty0 (1):\penalty0
	266--298, 2010.
	
	\bibitem[Hastie and Tibshirani(2000)]{hastie2000bayesian}
	Trevor Hastie and Robert Tibshirani.
	\newblock Bayesian backfitting (with comments and a rejoinder by the authors.
	\newblock \emph{Statistical Science}, 15\penalty0 (3):\penalty0 196--223, 2000.
	
	\bibitem[K{\"u}nzel et~al.(2019)K{\"u}nzel, Sekhon, Bickel, and
	Yu]{kunzel2019metalearners}
	S{\"o}ren~R K{\"u}nzel, Jasjeet~S Sekhon, Peter~J Bickel, and Bin Yu.
	\newblock Metalearners for estimating heterogeneous treatment effects using
	machine learning.
	\newblock \emph{Proceedings of the national academy of sciences}, 116\penalty0
	(10):\penalty0 4156--4165, 2019.
	
	\bibitem[Montgomery et~al.(2021)Montgomery, Peck, and
	Vining]{montgomery2021introduction}
	Douglas~C Montgomery, Elizabeth~A Peck, and G~Geoffrey Vining.
	\newblock \emph{Introduction to linear regression analysis}.
	\newblock John Wiley \& Sons, 2021.
	
	\bibitem[Breiman(2001)]{breiman2001random}
	Leo Breiman.
	\newblock Random forests.
	\newblock \emph{Machine learning}, 45\penalty0 (1):\penalty0 5--32, 2001.
	
	\bibitem[Aha(2013)]{aha2013lazy}
	David~W Aha.
	\newblock \emph{Lazy learning}.
	\newblock Springer Science \& Business Media, 2013.
	
	\bibitem[Alaa and Schaar(2018)]{alaa2018limits}
	Ahmed Alaa and Mihaela Schaar.
	\newblock Limits of estimating heterogeneous treatment effects: Guidelines for
	practical algorithm design.
	\newblock In \emph{International Conference on Machine Learning}, pages
	129--138. PMLR, 2018.
	
	\bibitem[Wager and Athey(2018)]{wager2018estimation}
	Stefan Wager and Susan Athey.
	\newblock Estimation and inference of heterogeneous treatment effects using
	random forests.
	\newblock \emph{Journal of the American Statistical Association}, 113\penalty0
	(523):\penalty0 1228--1242, 2018.
	
	\bibitem[Livieris et~al.(2018)Livieris, Kanavos, Vonitsanos, Kiriakidou,
	Vikatos, Giotopoulos, and Tampakas]{livieris2018performance}
	Ioannis~E Livieris, Andreas Kanavos, Gerasimos Vonitsanos, Niki Kiriakidou,
	Anastasios Vikatos, Konstantinos Giotopoulos, and Vassilis Tampakas.
	\newblock Performance evaluation of an ssl algorithm for forecasting the dow
	jones index stocks.
	\newblock In \emph{2018 9th International Conference on Information,
		Intelligence, Systems and Applications (IISA)}, pages 1--8. IEEE, 2018.
	
	\bibitem[Livieris(2020)]{livieris2020advanced}
	Ioannis~E Livieris.
	\newblock An advanced active set l-bfgs algorithm for training
	weight-constrained neural networks.
	\newblock \emph{Neural Computing and Applications}, 32\penalty0 (11):\penalty0
	6669--6684, 2020.
	
	\bibitem[Livieris and Pintelas(2019)]{livieris2019adaptive}
	Ioannis~E Livieris and Panagiotis Pintelas.
	\newblock An adaptive nonmonotone active set--weight constrained--neural
	network training algorithm.
	\newblock \emph{Neurocomputing}, 360:\penalty0 294--303, 2019.
	
	\bibitem[Friedman(1940)]{friedman1940comparison}
	Milton Friedman.
	\newblock A comparison of alternative tests of significance for the problem of
	m rankings.
	\newblock \emph{The Annals of Mathematical Statistics}, 11\penalty0
	(1):\penalty0 86--92, 1940.
	
	\bibitem[Garc{\'\i}a et~al.(2010)Garc{\'\i}a, Fern{\'a}ndez, Luengo, and
	Herrera]{garcia2010advanced}
	Salvador Garc{\'\i}a, Alberto Fern{\'a}ndez, Juli{\'a}n Luengo, and Francisco
	Herrera.
	\newblock Advanced nonparametric tests for multiple comparisons in the design
	of experiments in computational intelligence and data mining: Experimental
	analysis of power.
	\newblock \emph{Information sciences}, 180\penalty0 (10):\penalty0 2044--2064,
	2010.
	
	\bibitem[Bergmann and Hommel(1988)]{bergmann1988improvements}
	Beate Bergmann and Gerhard Hommel.
	\newblock Improvements of general multiple test procedures for redundant
	systems of hypotheses.
	\newblock In \emph{Multiple Hypothesenpr{\"u}fung/Multiple Hypotheses Testing},
	pages 100--115. Springer, 1988.
	
	\bibitem[Garcia and Herrera(2008)]{garcia2008extension}
	Salvador Garcia and Francisco Herrera.
	\newblock An extension on ``statistical comparisons of classifiers over
	multiple data sets'' for all pairwise comparisons.
	\newblock \emph{Journal of machine learning research}, 9\penalty0 (12), 2008.
	
	\bibitem[Wright(1992)]{wright1992adjusted}
	S~Paul Wright.
	\newblock Adjusted p-values for simultaneous inference.
	\newblock \emph{Biometrics}, pages 1005--1013, 1992.
	
	\bibitem[Hill(2011)]{hill2011bayesian}
	Jennifer~L Hill.
	\newblock Bayesian nonparametric modeling for causal inference.
	\newblock \emph{Journal of Computational and Graphical Statistics}, 20\penalty0
	(1):\penalty0 217--240, 2011.
	
	\bibitem[Dorie(2016)]{dorie2016npci}
	Vincent Dorie.
	\newblock Npci: Non-parametrics for causal inference.
	\newblock \emph{URL: https://github. com/vdorie/npci}, 2016.
	
	\bibitem[Gulli and Pal(2017)]{gulli2017deep}
	Antonio Gulli and Sujit Pal.
	\newblock \emph{Deep learning with Keras}.
	\newblock Packt Publishing Ltd, 2017.
	
\end{thebibliography}

\end{document}